# THE VERIFICATION-VALUE PARADOX: A NORMATIVE CRITIQUE OF GEN AI USE IN LEGAL PRACTICE

Joshua Yuvaraj[*]

It is often claimed that machine learning-based generative AI products will drastically streamline and reduce the cost of legal practice. This enthusiasm assumes lawyers can effectively manage AI's risks. Cases in Australia and elsewhere in which lawyers have been reprimanded for submitting inaccurate AI-generated content to courts suggest this paradigm must be revisited. This paper argues that a new paradigm is needed to evaluate AI use in practice, given (a) AI's disconnection from reality and its lack of transparency, and (b) lawyers' paramount duties like honesty, integrity, and not to mislead the court. It presents an alternative model of AI use in practice that more holistically reflects these features (the verification-value paradox). That paradox suggests increases in efficiency from AI use in legal practice will be met by a correspondingly greater imperative to manually verify any outputs of that use, rendering the net value of AI use often negligible to lawyers. The paper then sets out the paradox's implications for legal practice and legal education, including for AI use but also the values that the paradox suggests should undergird legal practice: fidelity to the truth and civic responsibility.

---

[*] Senior Lecturer, Faculty of Law, University of Auckland; Senior Fellow, Melbourne Law School; Academic Fellow, Centre for Technology, Robotics, Artificial Intelligence and the Law (TRAIL), National University of Singapore; Co-Director, New Zealand Centre for Intellectual Property. Email: Joshua.yuvaraj@auckland.ac.nz. Thanks to Jodi Gardner, Katy Barnett, Matt Bartlett, Julia Tolmie, Lucy Holloway, Andrew Godwin and the two peer reviewers for helpful comments on earlier drafts. Thanks also to the Law and Technology class at Auckland Law School, with whom this paper was discussed in a guest lecture. **Disclosure**: the author is a paid contributor to a commercial law looseleaf for Thomson Reuters, but is not involved in the development of any Thomson Reuters AI tools. **This is the draft of an article that has been accepted for publication in the** *Monash University Law Review*. **The final version of the article is forthcoming.**

# I INTRODUCTION

No shortage of commentary exists about the application of generative artificial intelligence (AI) to the legal profession across the world. That discourse is often positive: that AI represents a vast opportunity to streamline practices like legal research, document review, discovery, due diligence, the drafting of affidavits, submissions, contracts and other documents, and data analysis. The obvious appeal to law firm partners and executives, and barristers, is the potential to significantly reduce the time and cost associated with these tasks. There are also arguments that AI could be more accurate than humans at certain tasks, increasing the quality of the legal products and services offered to clients. This rhetoric explains a finding that of a representative sample of the world's law firms, over half had 'embraced AI technologies in some form' as of 2023.[1]

Such rhetoric is largely grounded in the view that AI provides great opportunity for lawyers, and that its risks can be effectively managed. However, the increasing number of cases where judges have criticised lawyers for filing hallucinated AI-generated material with courts calls this paradigm into question. In **Part I**I, this paper suggests that paradigm inadequately accounts for the impact of AI's structural flaws on legal practice. It then presents a different model to undergird evaluations of AI in legal practice in **Part III** (*verification-value paradox*). This model is grounded in AI's structural flaws, the nature of practice, and the paramount duties placed on lawyers as to honesty, integrity, the court, and the administration of justice. It suggests the value proposition of AI for legal practice is often overstated, and that in fact its net value is often likely to be negligible for much of the key work lawyers regard it as streamlining. It justifies this claim by reference to the *cost* and *imperative* of verification in **Part IV**. In **Part V**, it presents implications for this finding for legal practice and legal education.

The following caveats apply. First, while this paper adopts a broad, cross-jurisdictional conception of legal practice,[2] it is most closely grounded in approaches to lawyer regulation of common law jurisdictions like Australia, New Zealand, England and Wales, Singapore, and the United States. Second, this paper does not evaluate AI applications outside legal practice. For instance, AI may have utility in enhancing access to justice via online dispute resolution, precisely by allowing parties to bypass the traditional costs associated with engaging lawyers.[3] AI may also have some utility in judicial contexts, for example by making sentencing more consistent.[4] The model is designed to reflect the specific obligations and structure associated with legal practice.

---

[1] Michał Jackowski and Michał Araszkiewicz (eds), 'First Global Report on the State of Artificial Intelligence in Legal Practice', LLI White Paper (Liquid Legal Institute, 2023) 46.
[2] See a similar approach in Geoffrey C. Hazard, Jr and Angelo Dondi, *Legal Ethics: A Comparative Study* (Stanford University Press, 2004).
[3] See e.g. Vivi Tan, Jeannie Marie Paterson and Julian Webb, 'Generative AI in Small Value Consumer Disputes: Reviving Not Resolving Challenges of Design and Governance in Online Dispute Resolution' (2025) 48(4) *UNSW Law Journal* (forthcoming) <https://papers.ssrn.com/sol3/papers.cfm?abstract_id=5313052>; Wensdai Brooks, 'Artificial Bias: The Ethical Concerns of AI-Driven Dispute Resolution in Family Matters' (2022) 2022(2) *Journal of Dispute Resolution* 117.
[4] See e.g. Armin Alimardani and Milda Istiqomah, 'Beyond black boxes and biases: advancing artificial intelligence in sentencing' (2025) *Current Issues in Criminal Justice* 1; Richard Niall, 'AI and the Judicial System', Sir Ninian Stephen Law Program Oration 2025 (9 September 2025) 11 <https://www.supremecourt.vic.gov.au/sites/default/files/2025-09/Sir%20Ninian%20Stephen%20Oration%20-%20Chief%20Justice%20Niall%20-%20AI%20%20the%20Judicial%20System_0.pdf#page=11.26>.

Having said that, it is likely that the model reflects concerns about, and therefore can usefully inform discourse about AI use in industries like consulting, education, medicine, finance, and academic research.[5]

Third, this paper focuses on *generative*, rather than *predictive*, AI systems. Generative models are 'task-agnostic' systems trained to produce content (text, images, audio, video, etc) in response to user instructions.[6] This paper adopts the definition used by Tan, Paterson and Webb in their paper about the use of AI in small value consumer disputes:

> The term 'generative AI' describes AI models that can generate new outputs, such as text, image, video, or code. Generative AI is based on neural networks (a form of machine learning) using transformers, which learn from large volumes of data to create foundation models. A subset of foundation models, known as large language models ('LLMs'), are trained on textual data and use a technique known as natural language processing to interpret text-based prompts and generate textual responses. The main advance in accessibility with generative AI comes from an LLM being linked with a chatbot interface, which allows it to receive and respond to text (as opposed to code).[7]

Predictive models, however, are oriented towards predicting particular outcomes based on specific training data. Predictive analytics and predictive coding have been embedded in legal practice and discovery processes for some time,[8] and can have utility – for example, a sentence prediction model for criminal defence lawyers, the outputs of which are 'sense-checked' against and subordinate to the lawyer's experience and understanding.[9]

Given the relatively more integrated role of the latter category of AI into legal practice relative to the former, it is appropriate to maintain that dichotomy for the purposes of this paper, while acknowledging that generative models can be conceptualised as predictive (as set out below, the generative model 'predicts' which word/letter/pixel/other data point is likeliest to come next given its instructions and training data).[10] Whether, and how, the verification-value paradox engages predictive AI requires discrete engagement beyond the scope of this paper. However, given ongoing concerns about accuracy and bias in predictive analytics,[11] and the broad application of professional obligations to lawyers in all areas of their work, it is hypothesised that the model can readily be adapted to facilitate a holistic evaluation of those technologies in legal practice.

---

[5] See e.g. Paul Karp, 'Academics raise alarm over suspected AI use in Deloitte report', *Australian Financial Review* (online, 22 August 2025) <https://www.afr.com/companies/professional-services/academics-raise-alarm-over-suspected-ai-use-in-deloitte-report-20250822-p5mp0f>; Mike Thelwall et al, 'Does ChatGPT Ignore Article Retractions and Other Reliability Concerns?' (2025) 38(4) *Learned Publishing* e2018; Neil Selwyn, Marita Ljungqvist and Anders Sonesson, 'When the prompting stops: exploring teachers' work around the educational frailties of generative AI tools' (2025) 50(3) *Learning, Media and Technology* 310; Eimear Reynolds, 'Machine Learning-Integrated Medical Devices in Australia: Safety Defects and Regulation' (2025) 50(3) *Monash University Law Review* 467; Ross P. Buckley et al, 'Regulating Artificial Intelligence in Finance: Putting the Human in the Loop' (2021) 43(1) *Sydney Law Review* 43.
[6] See further Regulation (EU) 2024/1689 of the European Parliament and of the Council of 13 June 2024 laying down harmonised rules on artificial intelligence and amending Regulations (EC) No 300/2008, (EU) No 167/2013, (EU) No 168/2013, (EU) 2018/858, (EU) 2018/1139 and (EU) 2019/2144 and Directives 2014/90/EU, (EU) 2016/797 and (EU) 2020/1828 (Artificial Intelligence Act), Recital 99.
[7] Tan, Paterson and Webb (n 3) 4.
[8] Dan Hunter, 'The Death of the Legal Profession and the Future of Law' (2020) 43(4) *UNSW Law Journal* 1199, 1216-1218.
[9] See e.g. Harry Rodger, Andrew Lensen and Marcin Betkier, 'Explainable artificial intelligence for assault sentence prediction in New Zealand' (2023) 53(1) *Journal of the Royal Society of New Zealand* 133.
[10] The generative-predictive distinction is maintained elsewhere: see e.g. Jennifer Wang et al, 'Distinguishing Predictive and Generative AI in Regulation', arXiv:2506.17347v2 (2 July 2025) <https://doi.org/10.48550/arXiv.2506.17347>.

# II CHALLENGING THE RISK-OPPORTUNITY PARADIGM

## A *AI's structural flaws*

One consistent trend emerging from literature on AI and legal practice is that AI technologies present unique, new opportunities which lawyers must be equipped to effectively use in practice, but that such technologies come with significant risks those lawyers must be equipped to navigate.[12] This risk-opportunity paradigm allures lawyers, and future lawyers, with the possibility that they *can* reliably walk the tightrope to master the technology in service of greater efficiency and insights when providing legal services to clients.[13] That is a plausible-sounding explanation for the support for integrating AI into legal practice workflows. It also underpins the argument that legal educators *must* prepare law students to effectively use AI in practice.[14]

However, the paradigm fails to adequately engage structural flaws inherent to AI technology: failings of *reality* and *transparency*. These failings are common to all machine learning-based AI models that generate output based on instructions.[15] Until and unless a paradigmatic shift in technology occurs,[16] neither the scope nor sophistication of current or future AI models materially impacts the analysis below[17] - pushing back against the illusion that these deficiencies can simply be remedied. They are also structural features of both *general-purpose* AI services (that members of the public and/or organisations can use for any purpose, and purchase escalating tiers of service for, like ChatGPT, Microsoft Copilot, Gemini), *and* bespoke legal profession-targeted services (e.g. CoCounsel, Harvey AI). This distinction matters because claimed points of difference between the latter and former categories is their reliability and reliance on high-quality data for lawyers.[18] Such claims do not, however, address the fundamental structural flaws inherent to machine learning models outlined below.[19]

---

[11] John Zeleznikow, 'The benefits and dangers of using machine learning to support making legal predictions' (2023) 13 *WIREs Data Mining Knowledge Discovery* e1505, 12-13; see also Chris Chambers Goodman, 'AI/Esq.: Impacts of Artificial Intelligence in Lawyer-Client Relationships' (2019) 72(1) *Oklahoma Law Review* 149, 172-176.
[12] See e.g. Christy Ng, 'AI in the Legal Profession' in Larry A. DiMatteo, Cristina Poncibò and Michel Cannarsa (eds) *The Cambridge Handbook of Artificial Intelligence: Global Perspectives on Law and Ethics* (Cambridge University Press, 2022) 37; *Hussein v The Minister of Immigration, Refugees and Citizenship* 2025 FC 1138 [15].
[13] See e.g. Bruno Mascello, 'AI in the Legal Market: The Dark Side of the Moon' in Kai Jacob et al (eds) *Liquid Legal - Sustaining the Rule of Law: Artificial Intelligence, E-Justice, and the Cloud* (Springer, 2025) 97. See also an analogous comment about the underplaying of hallucinations: Susanne Förster and Yarden Skop, 'Between fact and fairy: tracing the hallucination metaphor in AI discourse' (2025) *AI & Society* 1, 12.
[14] Amanda Head and Sonya Willis, 'Assessing law students in a GenAI world to create knowledgeable future lawyers' (2024) 31(3) *International Journal of the Legal Profession* 293, 307.
[15] Burgess and Shareghi undertake a similar analysis of hallucination and explainability problems, but argue that they are fundamental problems for the practice of law. In this section I take a broader approach: these are structural problems not just for the practice of *law*, but *general* structural problems which affect the law among other disciplines and areas of society. Paul Burgess and Ehsan Shareghi, 'Not Explainable but Verifiable: Alternative First Steps in Overcoming the Problems Associated with AI's Answers to Legal Problems' in Phillip Hacker (ed) *Oxford Intersections: AI in Society* ('Why GenAI's Problems are Fundamental Problems (in Law)').
[16] See e.g. Samuel Becher and Benjamin Alarie, 'LexOptima: The promise of AI-enabled legal systems' (2025) 75(1) *University of Toronto Law Journal* 73.
[17] See Eliza Mik, 'Caveat Lector: Large Language Models in Legal Practice' (2024) 19(2) *Rutgers Business Law Review* 70, 121-123; Paul D. Callister, 'Generative AI and Finding the Law' (2025) 117 *Law Library Journal* 1, 49.

1 *The reality flaw*

AI models are fundamentally probabilistic[20] – they learn from input data, including its biases and omissions,[21] and return outputs that are statistically likeliest to reflect what is requested by users.[22] They are not *structurally* linked to reality: namely, factual accuracy, and valid links between 'factual propositions…[and] relevant legal documents.'[23] However accurate the training data, a machine learning model does not learn the facts *underlying* that training data, but reduces that data to patterns which it then ingests and seeks to reproduce with variations depending on the input of a user (instructions/prompts).

The reality flaw means hallucinations - outputs that are 'false, incorrect, or outright nonsensical', no matter how plausible-sounding[24] - and other errors occur frequently enough to warrant significant concern. One study of 'public-facing' models like GPT-4/3.5 (OpenAI), PaLM2 (Google) and Llama 2 found that, in response to 'a direct, verifiable question about a randomly selected [US] federal court case', 58%-88% of responses were hallucinations.[25] Similarly, a study of GPT-4o and Llama-3-8B legal analysis documented that 80% of responses had hallucinations.[26]

One response to this problem is to enhance the quality of the training data. In one sense this can help address concerns about omissions, biases, mistakes and other failings in the training data that can impact the output. Yet even with more high-quality training datasets, or bespoke datasets built for particular contexts, the hallucination problem is not immediately resolved. Studies have repeatedly found 'hallucinations' even in machine learning tools built for the legal context.[27] Even where leading legal research companies like Westlaw and Lexis have built AI into their search functions, they remain unreliable. In one study comparing the responses of ChatGPT-

---

[18] 'CoCounsel reveals the path towards trusted, transparent generative AI for business', *The Guardian* (online, 3 October 2024, paid advertising) <https://www.theguardian.com/thomson-reuters-ai-futures/2024/oct/03/cocounsel-reveals-the-path-towards-trusted-transparent-generative-ai-for-business>; Nicola Taljaard, 'Custom is the Future: How Harvey Lets Firms Build Their Own AI Systems', *The LegalWire* (online, 24 June 2025) <https://thelegalwire.ai/custom-is-the-future-how-harvey-lets-firms-build-their-own-ai-systems/>.
[19] See e.g. Brandon McHugh, David Myers and Ashvi Patel, 'AI Co-Counsel: An Attorney's Guide to Using Artificial Intelligence in the Practice of Law Symposium' (2024) 57(3) *Akron Law Review* 389, 394-395, where lawyer Brandon McHugh indicated their firm did not upload material to CoCounsel (despite it being a bespoke, purportedly secure AI system) due to concerns about privacy.
[20] Mik (n 17) 99.
[21] Alysia Blackham, 'Interrogating new methods in socio-legal studies: Content analysis, case law and artificial intelligence' (2025) 50(2) *Alternative Law Journal* 85, 89-90.
[22] Michael Legg, Vicki McNamara and Armin Alimardani, 'The Promise and the Peril of the Use of Generative Artificial Intelligence in Litigation' (2025) 48(4) *UNSW Law Journal* (forthcoming), 5.
[23] Varun Magesh et al, 'Hallucination-Free? Assessing the Reliability of Leading AI Legal Research Tools' (2025) 22 *Journal of Empirical Legal Studies* 216, 220-221.
[24] Mik (n 17) 93.
[25] Matthew Dahl, Varun Magesh, Mirac Suzgun and Daniel E. Ho., 'Large Legal Fictions: Profiling Legal Hallucinations in Large Language Models' (2024) 16 *Journal of Legal Analysis* 64, 66.
[26] Abe Bohan Hou et al, 'Gaps or Hallucinations? Scrutinizing Machine-Generated Legal Analysis for Fine-Grained Text Evaluations' (2024) *Proceedings of the Natural Legal Language Processing Workshop 2024* 280, 287.
[27] Josh Buckley et al, 'Towards a Legal Prompt Engineering Strategy for Identifying Rationes Decidendi' (2025) 51 *Monash University Law Review* 1, 21.

4, Copilot, DeepSeek, Llama 3 and Lexis+AI, the latter was the most accurate (58% 'precision'), but *still* returned 22% of responses incomplete, and 20% were hallucinations.[28] Meanwhile, another empirical study found Westlaw and Lexis's AI tools still have a hallucination rate of 17-33%.[29] These are astounding figures for tools specifically built to enhance legal research and *reduce* the hallucination problem that plagues general public-facing generative AI models, chiefly by ensuring higher-quality data from which outputs are drawn.[30]

The dangers of relying on even legal profession-specific models, or models purportedly trained on 'clean' legal datasets, are not hypothetical. In *Northbound Processing v South African Diamond Regulator*, for example, counsel for the plaintiff used an AI tool called 'Legal Genius' which purported to be trained only on South African legal materials, due to time pressures.[31] Because Northbound's submissions included 'incorrect citations',[32] the High Court of South Africa referred counsel to the South African Legal Practice Council, indicating it considered this behaviour potentially 'serious professional misconduct', even if negligent rather than intentional.[33]

The fact that all of the above products are likely to have undergone extensive post-training to minimise the risk of hallucinations and errors, including the use of human feedback,[34] only serves to emphasise how difficult it is to address the reality problem, let alone to the degree of accuracy required by lawyers. Accordingly, any output generated by AI must be verified if the user wishes to satisfy themselves as to the accuracy, and connection to reality, of that output – especially in legal practice.

## 2. *The transparency flaw*

The reality flaw is augmented by the *transparency* flaw. Software products, from operating systems to programs to videogames, are defined by the ability of programmers to review all underlying code. This is typically to identify and remedy bugs or other undesired outcomes: readers will instinctively be aware of this process by the frequency of application updates sent to their smartphones. In this context the software's 'decisions' can be scrutinised.

Machine learning models are fundamentally different because they operate as 'black boxes': while the parameters of a machine learning model and its training data can be set up in advance, how the model applies those parameters to that data in response to each instruction cannot be scrutinised. The inability to identify how decisions are made can lead to a reduction in trust.[35] This failing is not necessarily remediable by the release of parameter code or 'open source models' by AI companies[36] – those enable developers to manipulate *parameters* but not to evaluate the *application* of those parameters. Nor is it remediable necessary by 'reasoning' AI models, which 'generate a 'chain of thought' before responding to user queries'; that 'reasoning' may well be 'faked' in the sense that it does not accurately represent the processes the system underwent to reach that conclusion, but is presented as such[37] – though this does not imply any intention to mislead by the company or AI service provider. In fact, the lack of transparency, or 'opacity', appears structural, going to the heart of machine learning.[38]

---

[28] Bakht Munir et al, 'Evaluating AI in Legal Operations: A Comparative Analysis of Accuracy, Completeness, and Hallucinations in ChatGPT 4, Copilot, DeepSeek, Lexis+ AI, and Llama 3' (2025) *International Journal of Legal Information* 1, 8.
[29] Magesh et al (n 23), 216-217.
[30] For such discourse, see e.g. Frank Fagan, 'A View of How Language Models Will Transform Law' (2024) 92(1) *Tennessee Law Review* 1, 28; Elizabeth Chan, Kiran Nasir Gore and Eliza Jiang, 'Harnessing Artificial Intelligence in International Arbitration Practice' (2023) 16(2) *Contemporary Asia Arbitration Journal* 263, 288; Paul D. Callister, 'Generative AI and Finding the Law' (2025) 117 *Law Library Journal* 1, 11. See also Lee F. Peoples, 'Artificial Intelligence and Legal Analysis: Implications for Legal Education and the Profession' (2025) 117(1) *Law Library Journal* 52 [76]-[77], reporting findings that general purpose AI models actually 'outperformed Lexis+AI at legal analysis and reasoning.
[31] *Northbound Processing v South African Diamond Register*, High Court of South Africa, Gauteng Division, 2025-072038 [89].
[32] Ibid [86]-[88].
[33] Ibid [94]-[96].
[34] Lei Huang et al, 'A Survey on Hallucination in Large Language Models: Principles, Taxonomy, Challenges, and Open Questions' (2025) 43(2) *ACM Transactions on Information Systems* [2.2.3]

Whether the transparency problem can be overcome is presently unclear. A considerable body of scholarship exists around the concepts of 'Explainable machine learning' or 'Explainable artificial intelligence', which is designed to address the risk. Explainable AI (XAI) can be divided into global and local applications, or techniques.[39] The former is about 'the entire model's rationale, providing a comprehensive overview of the decision-making process and its various potential results', while the latter is about 'explaining individual decisions or predictions.'[40]

If the application of AI processes can be understood by humans, it is likely that the acceptability of AI use will increase in society generally and in legal practice specifically, not simply from an *ethical* perspective,[41] but from a utilitarian *cost* perspective because trust in the reliability of outputs will increase.[42] Yoo's broad conception of transparency (comprising disclosure of information about the training data, the code underlying algorithms, model testing and optimisation, and evaluation after release) is the type of framework likely to contribute to such increased public trust.[43]

However, it appears that effective XAI is still some way off. While there are some indications of limited increases in performance by users relative to 'black box' models,[44] there remains a lack of clarity about what terms like 'explainability and 'interpretability' mean in the machine learning literature, let alone what it means to functionally 'understand an AI system'.[45] Additionally, the different techniques that make up XAI are a live issue.[46] Until these issues are resolved, it appears no reliable solution presently exists for the transparency problem.[47]

---

[35] Emanuele Ratti and Mark Graves, 'Explainable machine learning practices: opening another black box for reliable medical AI' (2022) 2 *AI and Ethics* 801, 802.
[36] Ben Dickson, 'OpenAI's grand return to open source: unpacking the gpt-oss release', *TechTalks* (online, 5 August 2025) <https://bdtechtalks.com/2025/08/05/openai-gpt-oss-open-source-llm/>.
[37] Legg, McNamara and Alimardani (n 22) 6.
[38] Tatiana Dancy and Monika Zalnieriute, 'AI and Transparency in Judicial Decision-Making' (2025) *Oxford Journal of Legal Studies* (forthcoming), 31, citation omitted.
[39] Arthur Thu and Dries F. Benoit, 'Explainability through uncertainty: Trustworthy decision-making with neural networks' (2024) 317 *European Journal of Operational Research* 330, 331.
[40] Siru Liu et al, 'Leveraging explainable artificial intelligence to optimise clinical decision support' (2024) 31(4) *Journal of the American Medical Informatics Association* 968, 969.
[41] Timo Speith et al, 'Conceptualizing understanding in explainable artificial intelligence (XAI): an abilities-based approach' (2024) 26 *Ethics and Information Technology* 40, 40.
[42] On increased trust, see e.g. Regina de Brito Duarte, 'AI Trust: Can Explainable AI Enhance Warranted Trust?' (2023) *Hindawi: Human Behavior and Emerging Technologies* 4637678.
[43] Christopher S. Yoo, 'Beyond Algorithmic Disclosure for AI' (2024) 25 *Columbia Science & Technology Law Review* 314. See also Jennifer Cobbe and Jatinder Singh, 'Reviewable Automated Decision-Making' (2020) 39 *Computer Law & Security Review* 105475, 2 and Jennifer Cobbe and Jatinder Singh, 'Accounting for context in AI technologies' in Regine Paul, Emma Carmel and Jennifer Cobbe (eds) *Handbook on Public Policy and Artificial Intelligence* (Edward Elgar, 2024) for alternative models of transparency in automated decision-making.
[44] Julian Senoner et al, 'Explainable AI improves task performance in human-AI collaboration' (2024) 14 *Scientific Reports* 31150.

This is problematic because the need to verify outputs given the reality flaw is commensurate with the need to be able to explain how a particular output was generated.[48] The law may also require decisions to be explainable to allow external auditing of automated decision-making.[49] Yet as Hacker et al suggest, the complexities of emerging models may render it impossible to generate an explanation for a decision from an AI model, especially given 'not even experts can easily explain the outcome due to the huge number of parameters involved.'[50] There is an inbuilt tension here that does not present an easy resolution: the explainability of AI outputs is held in high regard by legal professionals like judges,[51] yet no clear technological path to explainability appears.

## B *Need for a different framework to evaluate AI in legal practice and education*

The risk-opportunity paradigm undergirds recommendations for the widespread integration of AI in legal practice because it assumes AI's risks can reliably be managed. Yet the reality and transparency flaws strongly suggest that paradigm is an inappropriate framework through which to evaluate the use of AI in legal practice. However, the steadily increasing number of cases in which AI-generated material has been used, or appears to have been used, with hallucinations or otherwise incorrect information,[52] suggests many lawyers have underestimated these flaws, and/or that they have been underplayed by technology companies marketing these products generally and to lawyers specifically.

---

[45] Shier Nee Saw, Yet Yen Yan and Kwan Hoong Ng, 'Current status and future directions of explainable artificial intelligence in medical imaging' (2025) 183 *European Journal of Radiology* 111884, 1; see also Timo Speith et al, 'Conceptualizing understanding in explainable artificial intelligence (XAI): an abilities-based approach' (2024) 26 *Ethics and Information Technology* 40, 40 n 1; Speith et al, 41; Marek Pawlicki et al, 'Evaluating the necessity of the multiple metrics for assessing explainable AI: A critical examination' (2024) 602 *Neurocomputing* 128282.
[46] Thu and Benoit (n 39).
[47] For more on the in-process state of generative AI explainability, see e.g. Wojciech Samek, 'Explaining and Interpreting Generative AI' in Philip Hacker et al (eds) *The Oxford Handbook of the Foundations and Regulation of Generative AI* (Oxford University Press, 2025).
[47] Ibid 430.
[48] Mik (n 17) 103.
[49] Philipp Hacker et al, 'Explainable AI under contract and tort law: legal incentives and technical challenges' (2020) 28 *Artificial Intelligence and Law* 415, 429. See also Wojciech Samek, 'Explaining and Interpreting Generative AI' in Philip Hacker et al (eds) *The Oxford Handbook of the Foundations and Regulation of Generative AI* (Oxford University Press, 2025)
[50] Ibid 430.
[51] See e.g. Dancy and Zalnieriute (n 38) 16, 21, 39.
[52] Damien Charlotin, 'AI Hallucination Cases' (online, n.d., last accessed 17 August 2025) <https://www.damiencharlotin.com/hallucinations/>. While the database is the most comprehensive one available it has limitations: e.g. '[i]t does not track the (necessarily) wider universe of all fake citations or use of AI in court filings'. Further, the process of sourcing, coding and updating the data are not apparent from the website, nor are there corresponding publications which set out the methodology. However, links are provided to the raw judgment files in many cases, and the entire dataset is available to download. Thus, the resource can be taken as a general indicator of trends of lawyers being reprimanded for AI use, but should not be regarded as comprehensive.

There are many potential reasons for inadequately verified AI use in legal practice. Legg, McNamara and Alimardani suggest 'AI literacy, legal literacy, copy-paste practice, automation bias and verification drift' can explain such behaviour.[53] A flexible moral compass may also be a plausible explanation. Yoon, in an empirical study of lawyers' perception of ethical behaviour in Ontario, Canada, found lawyers who benefited from unethical behaviour 'were less likely to say other lawyers would perceive an ethical breach…than respondents harmed by the conduct'.[54] Meanwhile, Vaughan and Nokes's empirical study of environmental lawyers found lawyers were very willing to 'us[e] uncertainty in the law to their clients' advantage'.[55] Strikingly, the authors noted that '[w]hilst acting in the best interests of clients was prominent in our study, acting with integrity was not.'[56]

The 'large and statistically significant' difference[57] in the Ontario study could plausibly be extended to unverified AI use. Lawyers may be aware of their ethical and moral obligations to the court and client, but may be more willing to excuse it if they perceive it benefits them, for example by reducing expenses and time. Likewise, Vaughan and Nokes's finding that integrity was low on the list of lawyer priorities could explain a willingness among lawyers to use AI as a 'shortcut' without adequate verification, even if in doing so they are reckless rather than intentional about the inclusion of fabricated material.

Of course, such explanations require further empirical corroboration. More analysis is needed on what causes inadequate verification. More research is also needed in both computer science and law about potential technological solutions to these structural problems.[58] Further, these flaws have arisen in the context of litigation, when many lawyers operate purely transactional practices (mergers and acquisitions, business/land sale and purchases, etc). The integration of AI into so many practices worldwide suggests the risk-opportunity paradigm has been inculcated across both practice types, such that these criticisms are valid whether uses are in litigation of transactional work. However, more targeted empirical research to this effect would enhance the discourse.

In the meantime, and to complement those lines of inquiry, this paper provides a theoretical model to help lawyers think more holistically and critically about AI use in legal practice – both in transactional and dispute resolution work. This model is both a hypothesis for future empirical research into AI use in legal practice, *and* itself carries normative force for reform in legal practice and education. This is because it is grounded in an understanding of AI's inherent *structural* flaws and lawyers' paramount obligations to the court and the administration of justice.

---

[53] Legg, McNamara and Alimardani (n 22) 16.
[54] Albert Yoon, 'In the Eye of the Beholder: How Lawyers Perceive Legal Ethical Problems' (2025) 22 *Journal of Empirical Legal Studies* 345, 354-355.
[55] Steven Vaughan and Karen Nokes, 'Role morality in action? An empirical exploration of the professional ethics of practising environmental lawyers' (2025) *Legal Ethics* 1, 2.
[56] Ibid 22.
[57] Yoon (n 54) 355.
[58] See e.g. Burgess and Shareghi (n 15) ('Verifiable Agent: What It Is and How It Can Help').

# III THE VERIFICATION-VALUE PARADOX

As an alternative model for evaluating the use of AI in legal practice, this paper presents the *verification-value paradox*. Given law schools prepare students for entry into the profession, and therefore require, or should require, students to undertake legal analysis similar to that which they will undertake in practice, the model is broadly applicable to the use of AI by students in law school as well.

This paradox is not *novel* in the sense of presenting *new* knowledge: many scholars, practitioners and judges have articulated similar problems with AI and emphasised the importance of verifying AI outputs. Analogies can be drawn to observations from macroeconomics as to the stagnation of productivity *despite* technological gains, and as to the use of AI in academic research.[59] Nevertheless, the paradox in this paper usefully advances the scholarship because it synthesises and contextualises these concerns into a theoretical model with implications for legal practice and legal education.

The paradox is grounded in the following logic: the *net value* of an AI model in legal practice can only be assessed once the efficiency gain (savings on time, salary costs, firm resource costs, etc) is offset by the corresponding verification cost (cost to manually verify AI outputs for accuracy, completeness, relevance, etc).

$$N\ [net\ value] = EG\ [efficiency\ gain\ ] - V\ [verification\ cost]$$

**Figure 1: Verification-value paradox**

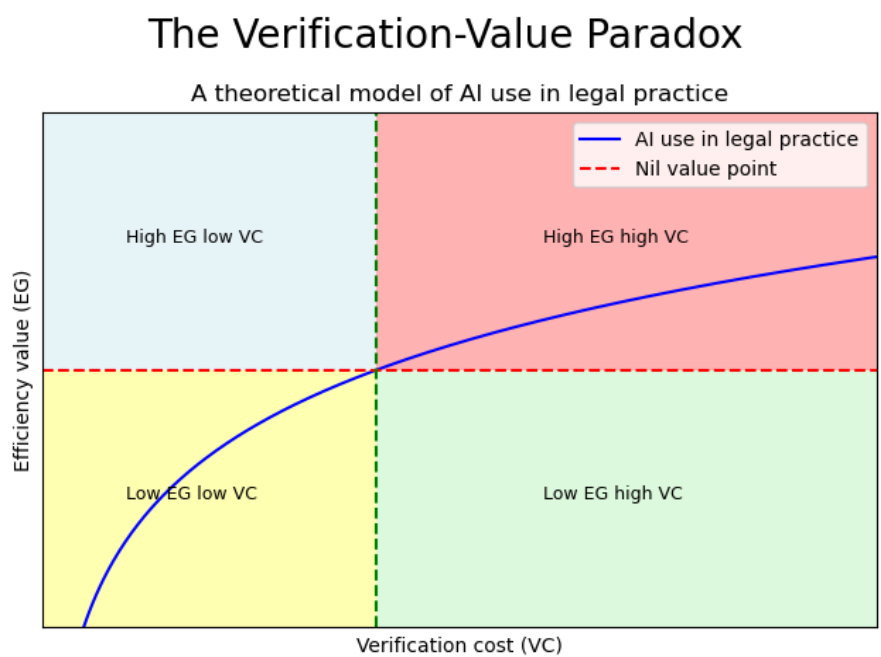

---

[59] Daron Acemoglu et al, 'Return of the Solow Paradox? IT, Productivity, and Employment in US Manufacturing' (2014) 104(5) *American Economic Review* 394, 399; Daniele Mezzadri, 'The Paradox of Ethical AI-Assisted Research' (2025) *Journal of Academic Ethics*.

## A *Different uses, different value*

1 *Bottom left quadrant: low efficiency gains, low verification cost*

This model highlights how, in legal practice, there are many uses which have low efficiency gains[60] and a correspondingly low verification cost (the **bottom left quadrant**). These can include locating and generating templates that would otherwise have taken time to identify in a firm's internal system. The outputs of such tasks can be relatively easily verified by quick visual checks precisely because their efficiency gain is low in that way: lawyers will still need to exercise considerable discretion and effort in populating such documents with advice, research, and supporting references. The verification requirement may also be lower if the work is 'nonlegal [or]…lower stakes…like drafting internal staff emails, social media content, policy talking points, policies, and administrative planning documents.'[61] The efficiency of such use cases may be enhanced if they can be deployed, with appropriate security for firm and client data, at reasonably low prices.

While potentially of some use, these efficiency gains arguably do not pass the *nil value point* – the point after which the efficiency gain compels lawyers to invest in AI as a substitute for, or at least in replacement of part of, a human workflow (even *before* the verification cost is taken into account). There are ongoing concerns that some of these types of AI uses in workplaces often produce low-quality filler, or 'AI slop', which can detrimentally affect productivity.[62] Moreover, they tend to fall into the 'nice-to-have' category when it comes to a legal workflow; of some use, but not generally the core tasks of the legal service, and therefore not worth shifting entire workflows towards AI.

2 *Top left/right quadrants: high efficiency gain, low/high verification cost*

Uses that may pass the nil value point likely include core tasks in respect of which it is argued that AI will save lawyers the most time, like document review, legal research, and drafting of submissions, affidavits, letters of advice, and other documents required by clients.[63] Such arguments may *appear* to be supported by empirical data. For instance, Schwarz et al found law students, when completing tasks like drafting client emails, legal memos, analysing complaints, drafting non-disclosure agreements, and drafting letters of advice, demonstrated 'statistically significant and meaningful improvements' for both the *quality* and *speed* of the work when using OpenAI's o1-preview (for general application) and VLex's Vincent AI model (specialised for lawyers).[64] The authors suggest that these models therefore 'enhance legal productivity and the quality of certain types of legal work.'[65] Another study of law students suggested adopting AI for legal tasks 'makes task completion about 30% faster, supporting the possibility that legal AI can make lawyers drastically more efficient without diminishing work quality.'[66]

---

[60] The same language is used in Colleen V. Chien and Miriam Kim, 'Generative AI and Legal Aid: Results from a Field Study and 100 Use Cases to Bridge the Access to Justice Gap' (2025) 57(4) *Loyola of Los Angeles Law Review* 903.
[61] Ibid 959.
[62] Kate Niederhoffer et al, 'AI-Generated "Workslop" Is Destroying Productivity', *Harvard Business Review* (26 September 2025) <https://hbr.org/2025/09/ai-generated-workslop-is-destroying-productivity?>; Jo Constantz, "AI Slop' Is Already Making a Mess in the Workplace', *Bloomberg* (online, 10 October 2025) <https://www.bloomberg.com/news/newsletters/2025-10-09/ai-slop-is-already-making-a-mess-in-the-workplace>.
[63] See e.g. Ralph Artigliere and Ralph C. Losey, 'The Future is Now: Why Trial Lawyers and Judges Should Embrace Generative AI Now and How to Do It Safely and Productively' (2025) 48(2) *American Journal of Trial Advocacy* 323, 324.

Meanwhile, interviews with English lawyers, in addition to secondary data, suggested machine learning – both extractive and generative – enabled lawyers to reallocate time away from more monotonous tasks (e.g. 'collating and summarising key facts and information…[and] producing boilerplate contractual/legal structures') and towards review, refinement, and decision-making.[67] And in a study of legal aid lawyers in California given access to generative AI programs like CoCounsel, GPT-4 and Gavel, over 90% of respondents reported some increased productivity, with 25% reporting significant or medium productivity increases.[68]

However, these findings paint an incomplete picture for the following reasons. First, inferences about these tools derived from observations of law students are not necessarily generalisable to lawyers, as integration of AI into legal practice might lead to different efficiency findings.[69] Second, the analyses of purported efficiency gains from the studies of students, and the experiences of lawyers whose workflows were augmented by machine learning, do not adequately account for the verification imperative: under the paradox, it calculates the y-values without the x-values. This omission can be seen in Schwarz et al's engagement with hallucinations – they noted those hallucinations were rare,[70] without engaging the impact of those hallucinations on the obligation to verify all content. Meanwhile, Chien and Kim's study of legal aid lawyers in California acknowledged the risks of hallucination to document summarisation, but focused anyway on the respondent-reported efficiency gains without considering the impact of verification.[71] One practical manifestation is that legal research will likely not be as effective: law arguably cannot be reduced to language, and doing so when using AI for legal research is likely to lead to inaccuracies at best and misleading outputs at worst.[72]

Accordingly, it is important to understand how efficiency gains correlate to the verification cost to effectively establish the net value of AI use in the profession. As Bell CJ of the New South Wales Supreme Court has commented extrajudicially, 'much more than efficiency gains are in play.'[73] The paradox hypothesises AI's purported efficiency gains will begin to taper off for legal

---

[64] Daniel Schwarz et al, 'AI-Powered Lawyering: AI Reasoning Models, Retrieval Augmented Generation, and the Future of Legal Practice', *Minnesota Legal Studies Research Paper No. 25-16, U of Michigan Public Law Research Paper No. 24-058* (4 March 2025), 30, 41 <https://papers.ssrn.com/sol3/papers.cfm?abstract_id=5162111>.
[65] Ibid 57.
[66] Aileen Nielsen et al, 'Building a better lawyer: Experimental evidence that artificial intelligence can increase legal work efficiency' (2024) 21 *Journal of Empirical Legal Studies* 979, 980.
[67] James Faulconbridge, 'Trajectories of legal work in the context of machine learning AI: conceptualizing mediated evolution' (2025) 32 *International Journal of the Legal Profession* 97, 106; see also Euan Black, 'Gen AI tools for lawyers 'hallucinate' up to one in three times', *Australian Financial Review* (online, 3 June 2024) <https://www.afr.com/work-and-careers/workplace/gen-ai-tools-for-lawyers-hallucinate-up-to-one-in-three-times-20240530-p5ji09>.
[68] Chien and Kim (n 60) 934-935.
[69] Nielsen et al (n 68) 1006-1007.
[70] Schwarz et al (n 66) 40.
[71] Chien and Kim (n 60) 942.
[72] Stefan Thiel, 'Carefully Tailored: Doctrinal Methods and Empirical Contributions' (2025) *Oxford Journal of Legal Studies* 1 (advance), 26-27.
[73] The Hon A S Bell, 'Change at the Bar and the Great Challenge of Gen AI', Address to the Australian Bar Association, 9 (online, 29 August 2025)

practice. Such reduced efficiency gains are consistent with observations that large language models are slowing in progress,[74] which challenges the law of AI scaling: 'feeding an AI system more data, expanding its model architecture, or boosting its computational power results in measurable performance gains.'[75] Further, as Narayanan and Kapoor note, scaling is an incomplete paradigm, failing to account for 'models' tendency to acquire new capabilities as size increases', which cannot be reliably measured.[76] They further note the potential limitation on these abilities in large language models because the evidence is not clear as to whether 'they [are]…capable of extrapolation or…only learn tasks represented in the training data'.[77]

While these observations are indicative, though, this paper makes no *conclusive* comment on the efficacy of technological mitigations for hallucinations at this point; only the general developmental slowdown of AI systems. Assuming such a slowing of efficiency gains, it is posited that any increase in that variable will be met by a greater verification cost; verification will take on even more importance, given the emphasis on accuracy and truth imposed on lawyers by professional practice standards and the law (discussed below). Thus, AI uses are likely to be in the **top right quadrant**,[78] suggesting that AI's net value in legal practice will be of much less utility than is often claimed.

The paradox of course requires empirical interrogation. However, there are indications that the paradox is borne out in reality. One empirical study of public-facing AI model responses to questions about US case law found 'hallucinations increase with the complexity of the legal research task at issue',[79] which reflects the reality flaw. Meanwhile, Carabantes argues that the more 'effective' a machine learning model is, the more opaque it is likely to be,[80] reflecting the transparency flaw. And Chien and Kim found that of lawyers who increased their use of AI, and reported greater productivity, 12.5% became 'more concerned' than they were prior to undertaking the study (with another 58% reporting being as concerned).[81] The breakdown of concerns is not provided, but included '[h]allucinations', '[e]thical concerns', and an '[i]nability to explain how AI

---

<https://supremecourt.nsw.gov.au/documents/Publications/Speeches/2025-speeches/bellcj/BellCJ-ABA-20250829.pdf>.

[74] Alex Wilkins, 'GPT-5's modest gains suggest AI's progress is slowing down', *New Scientist* (13 August 2025); Cal Newport, 'What if A.I. Doesn't Get Much Better Than This?', *The New Yorker* (online, 12 August 2025) <https://www.newyorker.com/culture/open-questions/what-if-ai-doesnt-get-much-better-than-this>; Kyle Wiggers, 'Improvements in 'reasoning' AI models may slow down soon, analysis finds', *TechCrunch* (online, 12 May 2025) <https://techcrunch.com/2025/05/12/improvements-in-reasoning-ai-models-may-slow-down-soon-analysis-finds/>; Deirdre Bosa and Jasmine Wu, 'The limits of intelligence – Why AI advancement could be slowing down', *CNBC* (online, 11 December 2024) <https://www.cnbc.com/2024/12/11/why-ai-advancement-could-be-slowing-down.html>; Melissa Heikkilä and Tim Bradshaw, 'The question suddenly sweeping through Silicon Valley', *Australian Financial Review* (online, 18 September 2025) <https://www.afr.com/technology/the-question-suddenly-sweeping-through-silicon-valley-20250818-p5mnrl>.

[75] Tao Hong and Ming Hu, 'Opportunities, Challenges, and Regulatory Responses to China's AI Computing Power Development under DeepSeek's Changing Landscape' (2025) *International Journal of Digital Law and Governance* 1, 4.

[76] Arvind Narayanan and Sayash Kapoor, 'AI scaling myths', *AI As Normal Technology* (Blog Post, 28 June 2024) <https://www.normaltech.ai/p/ai-scaling-myths>.

[77] Ibid.

[78] The formula is $EG = 2\log_8(VC)$. Microsoft Copilot was used to extract this formula from the code to the graph. The formula was then verified by discussions with Andrew Lensen and Andrew Chen, for whose input I am grateful.

[79] Dahl et al (n 25) 76.

[80] Manuel Carabantes, 'Black-box artificial intelligence: an epistemological and critical analysis' (2020) 35 *AI & Society* 309, 310, 313, 316.

[81] Chien and Kim (n 60) 934 n 155.

works',[82] which map to the reality and transparency flaws discussed above. It is plausible, then, to suggest that at least some of the studied lawyers became more anxious about the impact these flaws would have on their work as they saw how it purportedly increased their productivity – and therefore that verification of the results of that productivity became even more important.

To the extent that it is borne out in practice, then, the function suggests there are unlikely to be uses in the high-gain low verification cost quadrant which would lead to a positive net value justifying, or commending, AI's widespread adoption in legal practice (the **top left quadrant**). Rather, the verification cost will ultimately outperform any efficiency gain, which contrasts with the penchant to understate the verification cost and overstate AI's net value.[83] The nature of, and reasons for, high, and increasing relative to efficiency, verification costs in legal practice are discussed below.

# IV THE COST AND IMPERATIVE OF VERIFICATION

## A *The* cost *of verification in legal practice*

The underestimation of AI outputs can be partly explained by the fact that for visual, audio, and audiovisual outputs, verification is often instantaneous. For example, if an AI model is fed 1,000 images of black dogs that are labelled 'white cat', and generates an image of a black dog when asked to produce an image of a white cat, a human brain can typically assess in a fraction of a second that the image generated by the LLM is not that of a black dog. If, however, the model *had* been fed 1,000 correctly-labelled images of black dogs, and then produced an image of a black dog when so prompted, the net value of that LLM would be considerable because the corresponding cost of manual verification is miniscule – the visual check that all is as it seems.

The same logic applies to AI music and video generators, though the verification cost increases the longer the song or video. All of these uses could plausibly be placed in the **top-left quadrant**: they promise high efficiency gains by streamlining the process of generating music, images and/or video (as opposed to hours, or days, animating or producing such outputs), while the quality of those outputs can be verified visually or aurally in a fraction of the time it would have taken to create them. Similar reasoning applies to computer code.[84]

These examples show that AI use cases may well fit in the top left quadrant (high efficiency gain, low verification cost) in other contexts. However, purported uses in legal practice that involve significant efficiency gains are likely to be in the top right quadrant for their even greater verification costs, particularly for outward-facing products (analysis, statistics, advice, submissions to the court or clients). This is a corollary of the reality flaw: nothing generated by AI can be trusted in the first instance without verification, so the more important the output, the more important it is to verify its accuracy. It is also a corollary of the transparency flaw: with no reliable way to understand the reasoning of an AI model, external, manual verification becomes the only way by which practitioners can ensure the content they are presenting to clients, other practitioners and the courts are fully accurate.

---

[82] Ibid 977.
[83] See a similar observation as to academic research in Mezzadri (n 59) 6, and analogous comments about the effectiveness of AI at increasing productivity when it comes to 'hard' and 'easy' tasks: Daren Acemoglu, '*The Simple Macroeconomics of AI*' (2025) *Economic Policy* 13, 29-31.
[84] For example, **Figure 1** was generated with the assistance of Microsoft Copilot (GPT-5). The type of graph was explored by prompting Copilot and reviewing responses. Copilot was asked to provide the code in the Python programming language, which I used to independently generate the diagram and amended as appropriate to ensure the diagram reflected what was needed. The overall process took less time than it would have taken me to relearn Python coding skills I had long forgotten – but because I had those skills I could verify the code provided by Copilot was accurate and produced the diagram I sought, rather than simply trusting it. Verification was also necessary because the Copilot-generated code often could not run due to errors, or produced undesired outcomes. Considerable time was spent manually editing the code to achieve the final diagram. The curve function was also produced by Copilot, but verified by seeking advice from those with expertise in logarithmic functions.

As Kucharski argues, in certain use cases even a 1% risk of hallucinations is unacceptable.[85] Legal practice, it is submitted, falls within that category. Courts generally use a stringent, broad standard of verification of material presented to them: verification of AI-generated content in the legal context is not simply about whether sources cited *exist*, but whether claims are accurate, coherent, reasonably reflected in the cited source material.[86] In *May v Costaras*,[87] Bell CJ also indicated that in material filed with the court verification must include *existence* (all referenced material exists), *accuracy* ('references are accurate') and 'relevan[ce] to the proceedings.'[88] This conception applies in the context of AI-generated material in court proceedings, but there is no reason why a narrower approach should apply to the provision of any other legal services, unless one assumes clients deserve less scrutiny of advice provided to them than courts of submissions and affidavit evidence. As indicated in *Ayinde v London Borough of Haringey*, verification matters across the entire *spectrum* of legal work, not just court proceedings.[89] Thus, *all* legal work should require verification, broadly conceptualised, to ensure work generated using AI is 'trustworthy … encompass[ing] … metrics such as accuracy, consistency, and groundedness'.[90]

This conceptualisation of verification challenges arguments that verification could be automated.[91] For example, consider Burgess and Shareghi's 'verifiable agent' which would, if developed, 'autonomously interact with external law information repositories (e.g., AustLII, Casetext, or sources of legislation) to detect and retrieve relevant rules and facts from existing precedents and statutes…[so that] [t]he reasoning of the agent [is] grounded in these relevant references rather than the LLM's self-internal knowledge.'[92]

Such advancements are still unlikely to address reality and transparency flaws to the degree the profession requires. For instance, the risk of hallucinations remains considerable even when adopting automated verification processes.[93] Burgess and Shareghi also assume the hallucination problem is principally one of fictitious *authorities* for propositions of law, which their proposed verifiable agent could rectify by engaging with external repositories.[94] But as the Federal Circuit and Family Court indicates, that form of verification is inadequate:

> It is not sufficient to simply check that the cases cited were not fictious. What is expected from legal practitioners as part of their duty to the Court and to their client is that those cases (if they do exist) are reviewed to ensure they are authority for the principle the lawyer wishes to rely upon, have not been subsequently overturned or distinguished by a higher court, and are considered in respect of how and why those principles are relevant to the factual matrix of the case in which they intend to advance that proposition. Legal slogans are not simply slogans which can be affixed to submissions without context or analysis.[95]

---

[85] Adam Kucharski, 'Small hallucinations, big problems', *Understanding the unseen* (online, 18 August 2025) <https://kucharski.substack.com/p/small-hallucinations-big-problems>.
[86] See *JNE24 v Minister for Immigration and Citizenship* [2025] FedCFamC2G 1314 [34].
[87] *May v Costaras* [2025] NSWCA 178.
[88] Ibid [17].
[89] *Ayinde v The London Borough of Haringey* [2025] EWHC 1383 [7].
[90] Allison Koenecke et al, 'Tasks and Roles in Legal AI: Data Curation, Annotation, and Verification' (2025) *arXIV:2504.01349v1* [cs.CL] 9 <https://doi.org/10.48550/arXiv.2504.01349>.
[91] Buckley et al (n 27) 21. See also Bao Chau and Michael A. Livermore, 'Computational Legal Studies Comes of Age' (2024) 1(1) *European Journal of Empirical Legal Studies* 89, 95: 'e.g., training LLMs with adversarial examples designed to cause hallucination, applying post-processing filter to remove inconsistent outputs, and incorporating human feedback into the training process').
[92] Burgess and Shareghi (n 15) ('Verifiable Agent: What It Is and How It Can Help').
[93] Buckley et al (n 27) 33-34.

This high threshold of verification means it is still likely that lawyers will need to verify any AI output,[96] particularly in court proceedings where courts typically regard lawyers as entirely responsible for documents they submit.[97] The exact metrics of verification required will vary depending on the document and task for which AI is used. However, the principle remains the same: the more content AI generates for lawyers, the more that content must be manually verified. The verification-value paradox suggests that the more that needs to be manually verified, the less the efficiency gain. Accordingly, the net value of AI is unlikely to be enough to motivate widespread uptake in the legal profession.

## B *The* imperative *of verification in legal practice*

Verification is costly, which may encourage lawyers to short-circuit the process. Such an attitude is misguided for two reasons because verification is of the utmost importance when it comes to AI use in the law. While there are persuasive *moral* analyses as to the imperative of verification,[98] this paper focuses on the *regulatory* imperative: namely, the general rules governing practitioner conduct, specific judicial guidelines for AI use, and judicial criticism of practitioners who have presented hallucinated material to the courts. These points affect not just practitioners but law students, who are preparing to be subject to the profession's requirements. Thus, indicators of AI utility in university education in other disciplines like business[99] – to the extent they are valid – must be considered subject to these unique requirements.

While a comprehensive survey of all rules, judicial guidelines and commentary is beyond the scope of this paper, illustrative examples are adequate to highlight dominant perspectives on verification.

---

[94] Burgess and Shareghi (n 15) ('Overcoming Problem 1: Hallucinations'): '…the act of hallucinating (in a superficially very convincing way) creates authorities that do not exist'; ('Verifiable Agent: What It Is and How It Can Help').
[95] *JNE24* (n 91) [34].
[96] By analogy see Koenecke et al (n 95) 9 (on fact-checkers in legal writing) and Nicholas J. Abernethy, 'Let stochastic parrots squawk: why academic journals should allow large language models to coauthor articles' (2024) *AI and Ethics* 8 (on verification in an academic writing context).
[97] See e.g. *Ko v Li* 2025 ONSC 2965 [21].
[98] See e.g. Justine Rogers and Felicity Bell 'The Ethical AI Lawyer: What is Required of Lawyers When They Use Automated Systems?' (2019) 1 *Law, Technology and Humans* 80.
[99] See e.g. Alexander Richter and Ishara Sudeeptha, 'From fear to fluency: what our students learned when they used AI across an entire course', *The Conversation* (online, 26 August 2025) <https://theconversation.com/from-fear-to-fluency-what-our-students-learned-when-they-used-ai-across-an-entire-course-263805>.

1 *General rules governing practitioner conduct*

The potential damage caused by failing to appropriately verify information provided in the course of practice is highlighted by the seriousness with which rules governing lawyers emphasise honesty, integrity, fidelity to the court and the administration of justice. In Australia, the *Australian Solicitors' Conduct Rules* (ASCR) govern the behaviour of lawyers and are uniformly applied across all state jurisdictions. The ASCR provides that a lawyer's 'duty to the court and the administration of justice is paramount' to the exclusion of all other duties,[100] and that among their fundamental ethical duties are to 'be honest and courteous in all dealings in the course of legal practice', and 'to avoid any compromise to their integrity and professional independence.'[101]

Such an emphasis can be seen in the rules and principles governing solicitors in other common law countries. The New Zealand *Conduct and Client Care Rules* emphasises that those rules are 'minimum standards of professional conduct and client care…to maintain the reputation and integrity of the profession so as to ensure public confidence in the provision of legal services.'[102] Even so, lawyers 'acting in a professional capacity' must act 'with integrity, respect, and courtesy'.[103] In the UK, the Solicitors Regulation Authority imposes a paramount duty on lawyers to act with honesty and 'in a way that upholds public trust and confidence in the solicitors' profession'.[104] In Singapore, meanwhile, the *Legal Profession (Professional Conduct) Rules 2015* require lawyers to be 'honest in all the[ir]… dealings with [their]…client[s]', and must generally deal with 'any person[s]…[in an] honest and courteous [way]'.[105] These obligations are generally paramount in nature, and dishonesty is often grounds for removal from the profession.[106]

These general rules suggest that the obligation to verify the veracity and accuracy of all information provide in the course of legal services is paramount. An analogy for the seriousness of this obligation can also be drawn from corresponding obligations senior lawyers have to supervise more junior staff in the course of producing legal work. For example, the ASCR requires a supervising solicitor (those 'ultimately responsible for a client's matter or…responsible for supervising [that]…solicitor') to 'exercise reasonable supervision over solicitors and all other employees engaged in the provision of…legal services' in any particular matter.[107] The failure to supervise has been treated as professional misconduct in Australia.[108] The failure to verify content should therefore be considered in much the same way; in both categories, the lawyer fails to satisfy themselves as to the nature and/or requisite quality of the work they are presenting to the court or clients.

---

[100] Australian Solicitors Conduct Rules ('ASCR'), r 3.1 (online, November 2023) <https://lawcouncil.au/files/pdf/2023%20Nov%20-%20ASCR%20website%20version.pdf>.
[101] Ibid r 4.1.2, 4.1.4.
[102] *Lawyers and Conveyancers Act 2006* (NZ), Sch 1, r 1.5.1.
[103] Ibid Sch 1, r 12.
[104] 'SRA Principles', Solicitors Regulation Authority, para 1, 4 (n.d., online, last accessed 18 August 2025) <https://www.sra.org.uk/solicitors/standards-regulations/principles/>.
[105] *Legal Profession (Professional Conduct) Rules 2015* (Singapore), rr 5, 8(1)(a).
[106] See e.g. David J. Middleton, 'First Do No Harm, or Eat What You Kill? Why Dishonesty Matters Most for Lawyers' (2014) 17(3) *Legal Ethics* 382, 383, 386.
[107] ASCR, r 37.1.
[108] See e.g. *Law Society of New South Wales v Foreman* (1991) 24 NSWLR 238; *Victorian Legal Services Commissioner v Knight (Corrected) (Legal Practice)* [2025] VCAT 717 [111]-[114].

### 2 *GAI-specific guidelines for practitioners*

These general obligations have been applied to practitioners by courts and regulatory bodies *principally* in relation to conduct before the courts. For example, the Supreme Court of New South Wales requires that if AI is used for 'submissions…summaries or skeletons of argument, the author must verify… *that all citations, legal and academic authority and case law and legislative references… exist…are accurate…and…are relevant to the proceedings*', and do the same for 'references to evidence'.[109] The Victorian Supreme Court has issued similar guidelines, highlighting the role of practitioners 'to exercise judgment and professional skill in reviewing the final product to be provided to the Court', for issues such as currency, completion, accuracy, application to the Victorian jurisdiction, and bias in training data.[110]

Meanwhile, the Courts of New Zealand require '[a]ll information generated by a GenAI chatbot…[to] be checked by an appropriately qualified person for accuracy before it is used or referred to in court or tribunal proceedings.'[111] And courts in the United States have ordered that attorneys certify any material drafted using generative AI models has been thoroughly vetted, while Canadian courts have issued practice directions directing solicitors to disclose, and explain, the use of AI in the preparation of materials filed with the court.[112]

### 3. Judicial criticism

Courts have also reiterated the seriousness of the verification imperative where they have identified misleading, incomplete, irrelevant or inaccurate conduct placed before them that is either suspected or confirmed to have been generated by AI. Courts have emphasised two types of negative consequences of such content: professional and/or criminal liability for lawyers, threats to the administration of justice, and the potential damage to clients.

*(a) Potential damage for lawyers*

Failing to verify can have serious consequences for lawyers. This was emphasised by Dame Sharp DBE, President of the King's Bench Division of the High Court of England and Wales in *Ayinde*, in which non-existent cases or passages were submitted, or suspected to have been submitted to the Court.[113] These include police referrals for 'deliberately placing false material before the court with the intention of interfering with the administration of justice',[114] contempt of court findings for '[p]lacing false material before the court with the intention that the court treats it as genuine',[115] referrals to legal profession regulators,[116] costs orders against the solicitor for 'placing false material before the court with the intention of the court treating it as genuine…[being] improper and unreasonable and negligent conduct'.[117] Lawyers also risk public admonishment,[118] though Dame Sharp acknowledged that 'the risks posed to the administration of justice if fake material is placed before a court are such that, save in exceptional circumstances, admonishment alone is unlikely to be a sufficient response.'[119]

---

[109] New South Wales Supreme Court Practice Note SC Gen 23: Use of Generative Artificial Intelligence (Gen AI), 4-5 (online, 28 January 2025) <https://supremecourt.nsw.gov.au/documents/Practice-and-Procedure/Practice-Notes/general/current/PN_SC_Gen_23.pdf>.
[110] 'Guidelines for Litigants: Responsible Use of Artificial Intelligence in Litigation', Supreme Court of Victoria, para 8 (online, n.d., last accessed 3 September 2025) <https://www.supremecourt.vic.gov.au/forms-fees-and-services/forms-templates-and-guidelines/guideline-responsible-use-of-ai-in-litigation>.
[111] 'Guidelines for Use of Generative Artificial Intelligence in Courts and Tribunals', Courts of New Zealand (online, 7 December 2023) 4 <https://www.courtsofnz.govt.nz/assets/6-Going-to-Court/practice-directions/practice-guidelines/all-benches/20231207-GenAI-Guidelines-Lawyers.pdf>.
[112] Maura R. Grossman, Paul W. Grimm and Daniel G. Brown, 'Is disclosure and certification of the use of generative AI really necessary?' (2023) 107(2) *Judicature* 69, 70-71.
[113] *Ayinde* (n 94), cited with approval in *May v Costaras* (n 92) [13]; Robert Booth, 'High court tells UK lawyers to stop misuse of AI after fake case-law citations', *The Guardian* (online, 7 June 2025) <https://www.theguardian.com/technology/2025/jun/06/high-court-tells-uk-lawyers-to-urgently-stop-misuse-of-ai-in-legal-work>.
[114] *Ayinde* (n 94) [25].

The potential consequences elucidated by Dame Sharp are not illusory. Courts in Boston and Chicago have sanctioned lawyers for the use of ChatGPT to generate submissions involving fictitious cases and/or other material, including financial penalties.[120] The Federal Circuit and Family Court of Australia has also referred solicitors to legal practice regulators in Western Australia and Victoria for filing hallucinated material with inadequate verification, with the Victorian Legal Services Board varying the practitioner's practising certificate to prevent him from working as a principal lawyer, running his own practice or handling trust money, and requiring him to undertake supervised legal practice for two years with quarterly reports to the Board.[121] And as indicated above, even the use of a bespoke AI purportedly trained on South African legal resources was not enough to prevent hallucinations that led to an SC being referred to the South African Legal Practice Council.[122]

*(b) Potential damage to the profession and administration of justice*

The damage that can be done by inadequate verification is not just to lawyers, but to the profession and administration of justice. Dame Sharp highlighted this in *Ayinde*:

> As Dias J said when referring the case of Al-Haroun to this court, *the administration of justice depends on the court being able to rely without question on the integrity of those who appear before it and on their professionalism in only making submissions which can properly be supported*.[123]

These sentiments were echoed by the Supreme Court of Victoria in *Director of Public Prosecutions v GR*:

> The ability of the court to rely upon the accuracy of submissions made by counsel is fundamental to the administration of justice…any use of artificial intelligence without careful and attentive oversight of counsel would seriously undermine the court's processes and its ability to deliver justice in a timely and cost-effective manner.[124]

---

[115] Ibid [26].
[116] Ibid [29].
[117] Ibid [30].
[118] Ibid [31].
[119] Ibid.
[120] Lizzie Kane, 'Lawyer for CHA was sanctioned in previous case over AI use AI First incident involved hostile work environment lawsuit', *Chicago Tribune* (13 August 2025); Pat Murpy, 'Hallucinating' ChatGPT lands Boston lawyer in hot water', *Massachusetts Lawyers Weekly* (16 July 2025).
[121] *JNE24* (n 91) [22]; *Re Dayal* [2024] FedCFamC2F 1166 [19]-[22]; 'Statement on the 'Mr Dayal' matter', Victorian Legal Services Board + Commissioner (online, 2 September 2025) <https://lsbc.vic.gov.au/news-updates/news/statement-mr-dayal-matter>.
[122] *Northbound* (n 31) [96].
[123] *Ayinde* (n 94) [5], emphasis added.
[124] *Director of Public Prosecutions v GR* [2025] VSC 490 [79].

The Federal Circuit and Family Court in *JNE* went further in outlining exactly *how* inadequately verified AI use can negatively impact the administration of justice:

> There are now a concerning number of reported matters where reliance upon AI has directly led to the citation of fictitious cases in support of a legal principle. The dangers of such an approach are reasonably apparent but are worth stating. First, if discovered, there is the potential for a good case to be undermined by rank incompetence. Second, if undiscovered, there is the potential that the Court may be embarrassed and the administration of justice risks being compromised. Relatedly, the repetition of such cases in reported cases in turn feeds the cycle, and the possibility of a tranche of cases relying on a falsehood ensues. Further, the prevalence of this practice significantly wastes the time and resources of opposing parties and the Court. Finally, there is damage to the reputation of the profession when the clients of practitioners can genuinely feel aggrieved that they have paid for professional legal representation but received only the benefit of an amateurish and perfunctory online search.[125]

These are just some examples of how courts have specified verification is a serious obligation, which lawyers must comply with not merely to protect themselves, but to fulfil their broader duties to the administration of justice in both procedure (i.e. the timely and cost-effective resolution of disputes) and substance (i.e. the development of the law based on prior case law that exists). In this context it makes sense that verification must be 'independently and thoroughly verified.'[126] Indeed, the verification imperative is so great that the Ontario Court of Justice in *R v Chand* ordered defence counsel *not* to use AI to prepare submissions, given their initial submissions were marked by numerous apparent hallucinations.[127]

The administration of justice is not just an abstract goal; it speaks to the lived experiences of the parties who seek adjudication of their disputes, and who can be adversely affected by lawyer misconduct. The potential damage to clients should also factor into the assessment of the verification imperative. This additional element may explain why some judges have ordered lawyers found to have submitted AI-hallucinated material to inform their clients of the finding:[128] in a sense, it is forcing lawyers to confront head-on the people, companies, trusts or other entities to whom they owe duties, and whose cases they may have potentially jeopardised through the unsatisfactory conduct (at best) or negligence (at worst) of inadequately verifying material placed before the court.

All of the above suggests verification is not just costly; it is essential to the practice of law, especially in dealings with the court. Failing to discharge the verification imperative can have serious, adverse consequences for lawyers up to, and including, being disbarred and convicted. It can also have broader consequences for the administration of justice, including challenges to the integrity of the law itself to the extent that hallucinated material infects judgments in common law systems.[129]

---

[125] *JNE24* (n 91) [24].
[126] *GR* (n 131) [80].
[127] *R v Chand* 2025 ONCJ 282 [2]-[3], [5].
[128] See e.g. *Mata v Avianca*, USDC SDNY, 22-cv-1461 (PKC), Castel USDJ (22 June 2023) [33]; *Tajudin bin Gulam Rasul and anor v Suriaya bte Haja Mohideen* [2025] SGHCR 33 [100].
[129] See further Mike Scarcella, 'Two US judges withdraw rulings after attorneys question accuracy', *Reuters* (online, 30 July 2025) <https://www.reuters.com/legal/government/two-us-judges-withdraw-rulings-after-attorneys-question-accuracy-2025-07-29/>.

At this point one might potentially argue: what of the transactional and in-house lawyers who never engage with the court? Three brief comments suffice. First, Dame Sharp in *Ayinde* indicated that practitioners have a 'professional duty' to verify AI-generated content presented to the courts *or* as legal advice.[130] This strongly indicates that the same standard of verification applies to outputs regardless of location. Second, it would be illogical to suggest that lawyers may somehow be more lackadaisical with verification when providing client-facing work than court-facing work. Third, the courts are the ultimate arbiters of professional discipline in the legal profession, for example if regulatory decisions are challenged. It is again illogical to suggest they would view AI-generated omissions, mistakes or inaccuracies any less strictly for the purposes of disciplinary adjudication, if used in the process of advising clients than if submitted to the court. Accordingly, it is strongly arguable that the imperative of verification extends to *all* practitioners who use AI in the provision of *all* legal services, not just in their dealings with the court.

### C *Summarising the paradox*

The verification-value paradox suggests the net value of AI to legal practice is grossly overestimated, due to an underestimation of the verification cost. A proper understanding the costly and essential nature of verification leads to the conclusion that AI's net value will often be negligible in legal practice: that is, in most cases, the value added will not be sufficient to justify the corresponding verification cost. The next Part of this paper examines potential implications of this paradox for research, practice, and education.

## V IMPLICATIONS

The net value calculation remains sound because it is ultimately grounded in the paramount obligations lawyers have, which are unlikely to change with the same pace with which AI technology develops. However, the actual paradox – based on the above analysis of the verification cost – is fundamentally theoretical, even though anecdotal evidence seems to bear it out. The paradox is therefore a hypothesis for further examination.

The paradox and its implications could be revisited if data shows the verification cost changes; for example, as between generative AI uses and uses of machine learning for e-discovery or outcome prediction, or between AI uses for clients (drafting contracts, letters of advice) and uses for the court. New models built with elements like safety and ethics at their core,[131] to the extent they viably assist legal practice, may also be more trustworthy and therefore reduce the verification cost and lead to the re-categorisation of AI use closer to, or within, the top-left quadrant (high efficiency, low verification cost).

Until and unless such paradigmatic shifts occur, though, the paradox's most immediate implication is for lawyers to be cautious, critical and/or reticent as to AI use in legal practice. However, the paradox also has implications beyond facilitating efficiency calculations for AI use.

---

[130] *Ayinde* (n 94) [7].
[131] See e.g. Yoshua Bengio, 'Superintellligent Agents Pose Catastrophic Risks: Can Scientist AI Offer a Safer Path?' arXiv:2502.15657; Yoshua Bengio, 'A Potential Path to Safer AI Development', *TIME* (online, 9 May 2025) <https://time.com/7283507/safer-ai-development/>; Philip Brey and Brandt Dainow, 'Ethics by design for artificial intelligence' (2024) 4 *AI and Ethics* 1265.

It encourages lawyers to use the rise of AI to critically reflect on the values that *should* undergird legal practice.[132] Such reflections are not only for practitioners, but for law schools given they form the values of the next generation of practitioners.[133] In this broader context, the paradox suggests two values should be re-emphasised in legal practice and pedagogy: truth and civic responsibility.[134]

# A *Truth-centred practice and pedagogy*

While regulations and legal rules governing the conduct of lawyers highlight the primacy of honesty and integrity in legal practice, fidelity to the truth can often be relegated below other priorities like those of their clients.[135] The same can apply to law students, who may prioritise other values more highly.[136] However, the verification-value paradox lends itself to a truth-centred practice and pedagogy because it grounds itself on the fundamental value of the truth to the practice of law:[137] that is, that the output of a legal practitioner must be grounded in verifiable facts.[138]

## 1 *A truth-emphasis in legal practice*

In this context, a lawyer's fidelity to the truth draws on both consequentialist and deontological rationales. From a consequentialist lens, an adherence to the truth is essential to the administration of law and public confidence in governing institutions.[139] As the Federal Circuit and Family Court warns, undue reliance on AI can cause 'the repetition of [fictitious] cases in reported

---

[132] This paper does not directly respond to the question of what 'ethical' use of AI would look like more broadly. For more on 'practitioner' (including but not limited to legal practitioner) views on AI ethics (amongst the views of others like lawmakers), see e.g. Arif Ali Khan et al, 'AI Ethics: An Empirical Study on the Views of Practitioners and Lawmakers' (2023) 10(6) *IEEE Transactions on Computational Social Systems* 2971.

[133] For the purposes of this paper it is unnecessary to resolve the inherent tension between seeing legal education as a dedicated discipline and professional training, though that debate is acknowledged and ongoing in Australia: Daniel Goldsworthy, 'The Future of Legal Education in the 21st Century' (202) 41(1) *Adelaide Law Review* 244, 245-250.

[134] These implications complement those in Legg, McNamara and Alimardani (n 22) 24-34.

[135] Marvin E. Frankel, 'The Search for Truth: An Umpireal View' (1975) 123(5) *University of Pennsylvania Law Review* 1031, 1032; W. Bradley Wendel, 'Whose Truth: Objective and Subjective Perspectives on Truthfulness in Advocacy' (2016) 28(1) *Yale Journal of Law & the Humanities* 105, 110; Vaughan and Nokes (n 50) 22. A contrary position challenging the role of integrity in legal practice can be found in Daniel Markovits *A Modern Legal Ethics: Adversary Advocacy in a Democratic Age* (Princeton University Press, 2008) 135-136. In this paper, I refer to integrity as it pertains to solicitors' fidelity to the truth, though there are multiple definitions of integrity and it is sometimes thought of as 'denot[ing] a higher moral standard than honesty': Steven Vaughan, 'Existential Ethics: Thinking Hard About Lawyer Responsibility for Clients' Environmental Harms' (2023) 76 *Current Legal Problems* 1, 13-14.

[136] See e.g. Richard Wu and JaeWon Kim, 'An Empirical Study of Values of Law Students in South Korea: Does 'Americanized' Legal Education Impact Their Confucian Ethics?' (2022) 17 *University of Pennsylvania Asian Law Review* 209, 237.

[137] See further Frankel (n 142) 1055-1056.

[138] By analogy to journalism see e.g. Ana Azurmendi, 'Does It Still Make Sense to Talk About Journalistic Truth?' (2025) 40(2) *Journal of Media Ethics* 102,103, 108, 114. Of course, different practitioners may consider their 'truth' to be right to the exclusion of another's in an adversarial context. Further, lawyers may face difficult questions about the extent to which their obligation to maintain client confidentiality absolves them of the obligation to disclose to the court information that would be damaging to their client. These legitimate ethical conundrums are important but are outside the scope of this paper. For more, see W. Bradley Wendel, 'Whose Truth: Objective and Subjective Perspectives on Truthfulness in Advocacy' (2016) 28(1) *Yale Journal of Law & the Humanities* 105, 111; David Moss and Lance S. Bush, 'Measuring metaaesthetics: Challenges and ways forward' (2021) 62 *New Ideas in Psychology* 100866, 2; Douglas R. Richmond, 'Lawyers' Duty of Confidentiality and Clients' Crimes and Frauds' (2022) 38(2) *Georgia State University Law Review* 493.

[139] Kenneth Townsend, 'Purpose, Practical Wisdom, and the Formation of Trustworthy Lawyers' (2024) 75(5) *Mercer Law Review* 1399, 1399.

cases…feeds the cycle, and the possibility of a tranche of cases relying upon a falsehood ensues'.[140] Trustworthiness in lawyers is also essential for the smooth operation of markets and the resolution of disputes. Further, the consequences to lawyers of not upholding duties of integrity and honesty can be severe, both to them (disciplinary findings; criminal convictions) and others.

From a deontological perspective, truth matters because it is the truth: it is fundamental to humanity that we are creatures of truth, tethered to reality, designed to operate in consistency with and not in opposition to truth.[141] Indeed, deontological motivations can restrain behaviour where consequentialist ethics would place no hurdle to lying.[142] And those motivations generally reflect the paramount and qualifier-less nature of the obligations lawyers have towards honesty and integrity. Such deeper motivations should be encouraged to avoid situational ethics. The risk otherwise is that unverified AI use, as with other contraventions of honesty obligations in practice, becomes behaviour which lawyers are more willing to tolerate if it benefits them.[143]

A truth-centred approach to practice, incorporating both consequentialist and deontological motivations, will enable lawyers to properly appraise the value of truth to society as a whole, and therefore in their roles as officers of the court and in supporting the administration of justice.[144] For such individual decision-making by lawyers, Rogers and Bell's application of Rest's model of moral behaviour to the use of AI by lawyers is helpful: lawyers must be aware of the moral implications of their behaviour, determine the most morally appropriate response given their professional obligations, prioritise their moral values and decide to act on them, and then action their moral reasoning.[145] The analysis in this paper suggests applying this model will *typically* result in reticence towards integrating AI models in practice.

From a broader perspective, professional development/continuing legal education initiatives and lawyers' conferences can emphasise lawyers' broader obligations to courts and the administration of justice to reinforce the value of truth. Law societies can also foster mentoring relationships between senior and junior practitioners that provide alternative avenues for this emphasis outside the firm structure, reminding practitioners their fidelity lies not firstly to the client but to the administration of justice.

## 2 *A truth-emphasis in legal education*

The truth-based approach also suggests a critical attitude should be adopted to AI integration in legal education. This approach challenges Head and Willis's 'knowledge framework', which they use to ground recommendations to integrate AI into legal education:

> First, students require substantive legal knowledge in key areas (substantive legal knowledge). The ubiquity of GenAI has now imposed the second and third knowledge areas for optimal legal education, namely the knowledge of the legal and ethical risks of GenAI (GenAI ethics knowledge) and the skills to use GenAI effectively (GenAI system knowledge). Historically, engaging with any technology played a secondary role in legal education. This is arguably no longer a tenable position.[146]

---

[140] *JNE24* (n 91) [24].
[141] See further Harald Brüssow, 'What is truth – in science and beyond' (2022) 24(7) *Environmental Microbiology* 2895, 2905.
[142] Eberhard Feess, Florian Kerzenmacher and Yuriy Timofeyev, 'Utilitarian or deontological models of moral behavior – What predicts morally questionable decisions?' (2022) 149 *European Economic Review* 104264, 12.
[143] Yoon (n 54) 354-355.
[144] A similar argument is made about the role of ethical judgment in legal practice in Michael Legg, 'Better than a bot – instilling ethical judgement into the lawyers of the future in the age of AI' (2024) 33(3) *Griffith Law Review* 273, 279-283.
[145] Rogers and Bell (n 103) 86-94.

This framework is illusory because it grounds the normative recommendation in AI's 'ubiquity'. Yet ubiquity alone does not, and should not, determine what is taught to law students. Unduly inflating client invoices is also widespread in legal practice;[147] but no one would argue that law schools should equip students to do so. The influential factor is not *merely* the ubiquity of the practice. It is also whether the practice is consistent with the obligations imposed on lawyers.

The verification-value paradox does not suggest AI use is inherently contrary to those obligations. However, as courts have indicated, it *may* become contrary to those obligations if verification is inadequate. This challenges the argument that law students must be prepared for the expectations placed on them to use AI technology.[148] The question should instead be: what must law students know about AI to faithfully discharge their professional obligations to the administration of justice, the courts and their clients? The verification-value paradox suggests the answer is *not* how to 'use GenAI effectively'. It suggests AI use will, absent paradigmatic technological shifts nullifying reality and transparency flaws, always require external verification, largely negating AI's purported efficiency gains in practice. Accordingly, aspiring lawyers are likely to remain compliant with paramount duties, including the duty of competence,[149] *without* integrating AI into their workflows.

This is likely the case even where competence expressly requires technological competence, as set out in Comment 8 to the American Bar Association's Model Rule 1.1. That comment requires 'a lawyer…[to] keep abreast of changes in the law and its practice, including the benefits and risks associated with relevant technology…'[150] far short of a requirement to integrate AI into practice. This was reiterated in a 2024 Formal Opinion by the ABA's Standing Committee on Ethics and Professional Responsibility, which indicated that competent use of generative AI when representing clients does *not* require lawyers to 'become GAI experts… [but to] have a reasonable understanding of the capabilities and limitations of the specific GAI technology that the lawyer might use.'[151] This knowledge is consistent with a truth-centred pedagogy, but stops far short of *mandating* AI's integration into legal pedagogy.

---

[146] Head and Willis (n 14) 295-296.

[147] Christine Parker and David Ruschena, 'The Pressures of Billable Hours: Lessons from a Survey of Billing Practices Inside Law Firms' (2011) 9(2) *University of St. Thomas Law Journal* 619, 641-642.

[148] Head and Willis (n 14) 306-307; see also Marjan Ajevski et al, 'ChatGPT and the future of legal education and practice' (2023) 57(3) *The Law Teacher* 352, 363-364. Of course, law graduates often enter career pathways outside the law, for which AI may well be relevant and useful. However, this does not suggest law schools *should* train students for those professions, any more than they should train them in economic theory, statistics or philosophy. Such knowledge may well be incorporated where relevant to law papers; but the core pedagogical distinctive of the law degree remains *the law* and how to understand and wield it in advocacy for clients, whether in transactional or dispute contexts.

[149] See e.g. Australian Solicitors Conduct Rules (n 92) r 4.1.3; American Bar Association, Model Rule 1.1; *Lawyers and Conveyancers Act (Lawyers: Conduct and Client Care) Rules 2008* (NZ), r 3; SRA Code of Conduct for Solicitors, RELs, RFLs and RSLs, r 3.2, 3.3, 3.6 (online, n.d., last accessed 3 September 2025) <https://www.sra.org.uk/solicitors/standards-regulations/code-conduct-solicitors/#:~:text=You%20do%20not%20mislead%20or,or%20discriminate%20unfairly%20against%20them.>.

[150] American Bar Association, Rule 1.1 Competence – Comment (online, n.d., last accessed 3 September 2025) <https://www.americanbar.org/groups/professional_responsibility/publications/model_rules_of_professional_conduct/rule_1_1_competence/comment_on_rule_1_1/>.

In practice, law schools could integrate a truth-based pedagogy by increasing student awareness of these issues,[152] minimising the permitted use of AI technologies for assessments, and actively discouraging students from using AI as part of their learning.[153] This would likely mean refocusing assessment regimes away from those assessments in respect of which AI use cannot be monitored, like essays or take-home exams, towards more heavily-weighted, secure final examinations incorporating critical reflection, comprehension and rule application, and skills-based assessments, like oral presentations, mock trials, client negotiations, and more to supplement final exams.

This analysis may well require revisiting if there are paradigmatic shifts either to AI technology (e.g. the achievement of artificial general intelligence) or accepted standards of behaviour in the profession (e.g. an exemption from liability for practitioners who use AI to generate court submissions with *reasonable*, not *complete*, verification).[154] For example, in *McConnell Dowell Constructors v Santam (No 1)*, the Victorian Supreme Court endorsed the use of technology assisted review for the purpose of discovery management.[155] This system bears similarities to AI insofar as 'the software enables a computer to be 'trained' to recognise concepts in the electronic documents fed into the system which are relevant to the issues in the proceeding.'[156] The widespread acceptance of such technologies for discovery means there is a reasonable case that familiarity with those technologies will be strongly influential, if not determinative, on decisions to employ or retain young lawyers.

---

[151] American Bar Association Standing Committee on Ethics and Professional Responsibility, 'Formal Opinion 512: Generative Artificial Intelligence Tools', 2-3 (online, 29 July 2024) <https://www.americanbar.org/content/dam/aba/administrative/professional_responsibility/ethics-opinions/aba-formal-opinion-512.pdf#page=2.40>. This also appears to be the emphasis behind the call for bar entrance examinations to prioritise 'information literacy': Amy A. Emerson, 'Assessing Information Literacy in the Age of Generative AI: A Call to the National Conference of Bar Examiners' (2025) 44(1) *Legal Reference Services Quarterly* 41, 96.
[152] Head and Willis (n 14) 306-307. See also Stephanie L. Grace, 'Finding Equilibrium: An Integrative Approach to Balancing Human and Artificial Intelligence in Legal Research' (2025) *Legal Reference services Quarterly* 1, 32.
[153] See the case study conducted in Head and Willis (n 14), involving the provision of AI outputs to students and documenting responses: 299. Such a reflection may be useful to highlight AI's flaws, though Head and Willis had a different framework to the verification-value paradox: 295-296. For contrary positions, see e.g. Sara Migliorini and João Ilhão Moreira, 'The Case for Nurturing AI Literary in Law Schools' (2024) 12(1) *Asian Journal of Legal Education* 7; Stuart Hargreaves, "Words Are Flowing Out Like Endless Rain Into A Paper Cup': ChatGPT & Law School Assessments' (2023) 33 *Legal Education Review* 69, 90-93; Jack Wright Nelson, 'The 'other' LLM: large language models and the future of legal education'(2024) 5(1) *European Journal of Legal Education* 127.
[154] For example, in empirical research the reliability of coding instructions used to systematically analyse texts (judgments, legislation etc) is calculated by the level of agreement between two independent coders. Perfect agreement is not required, only agreement over a particular threshold using a statistical indicator. See e.g. Mark A. Hall and Ronald F. Wright, 'Systematic Content Analysis of Judicial Opinions' (2008) 96 *California Law Review* 63, 115-116. Another example is Davidov's proposal to use AI to 'pre-authoris[e]' an AI classification of a worker as an independent contractor, where the AI system 'predicts with at least 51 per cent certainty that the worker would be considered an independent contractor by the court', if that assessment became accepted by courts in employment law: Guy Davidov, 'Using AI to Mitigate the Employee Misclassification Problem' (2025) 88(2) *The Modern Law Review* 267, 282. There are some indications of this approach to verification in the American Bar Association's guidance for lawyers using GAI: American Bar Association (n 158) 4. However, this is guidance only, and stops short of a court-approved liability exemption for GAI errors.
[155] [2016] VSC 734.
[156] Ibid [20].

For the reasons outlined above, there are still structural issues to be addressed if generative AI systems are to reach the high level of automated verification required to merit similarly widespread acceptance in legal practice. However, even assuming that threshold *is* met, to the extent that bespoke AI expertise is 'required' for graduates, the intelligence and capability of most law students – given the rigour of law school admissions processes – *should* give some hope that they can be learned relatively quickly, either informally, through graduate supervision schemes or through dedicated, short practical legal training programs required for admission to practice.[157]

In jurisdictions where clinical legal training is required as part of qualifying legal education,[158] such skills can be developed there, rather than in substantive law papers. Doing so would address pressures from regulators, firms and other industry stakeholders to ensure new entrants to the profession are technologically competent, because they would be hurdle programs for admission to the profession. Far more fundamental than understanding how to apply these technologies, though, is for *law schools* to inculcate in students a deep appreciation for the value of truth, which will impact all of their practice, including how they incorporate technology. Doing so will combat the potential AI has 'to encourage or feed laziness in research and analysis and a loss of essential skills and critical thinking'.[159] A truth-centred pedagogy, rather than AI-literacy, is the key 'to produc[ing] a new generation of competent, knowledgeable lawyers';[160] or, in Bell CJ's words, 'to ensure that legal graduates have demonstrated that they have a genuine and personal understanding of fundamental legal principles.'[161]

## B *Emphasise civic responsibility*

The verification-value paradox should also encourage the development of civic responsibility in lawyers and law students: 'attitudes and behaviors that are beneficial to society…typically result[ing] from an interest to promote the common good.'[162] The cost of verification under the paradox is high precisely because the truth matters, and citizens, the judiciary, businesses and other members of society must be able to take lawyers at their word, written or verbal.

However, this is not simply because truth matters deontologically. Society relies on the trustworthiness of lawyers to function. For example, lawyers' undertakings are viewed as sacrosanct, and firm enough bases on which to transfer large sums of money and change land title arrangements. Lawyers are also trusted, alongside doctors, pharmacists and other limited professions, to certify documents and witness affidavits and statutory declarations.[163] These are just some illustrations of how 'lawyers are crucial participants in the development of the infrastructure of civil society'.[164] Lawyers must steward such power responsibly for the common good of society, in ways that 'transcend self-interest.'[165] This power explains the importance of effective regulation, and the consequences of not meeting those standards as articulated above (up to, and including, potential criminal liability for use of hallucinated materials in proceedings).

---

[157] See e.g. Maxine Evers, Bronwyn Olliffe and Robyn Pettit, 'Looking to the past to plan for the future: a decade of practical legal training' (2011) 45(1) *The Law Teacher* 18, 32.
[158] See e.g. American Bar Association, Standard 303(a).
[159] Bell (n 75) 33.
[160] Head and Willis (n 14) 307.
[161] Bell (n 75) 34.
[162] Lisa da Silva et al, 'Civic Responsibility Among Australian Adolescents: Testing Two Competing Models' (2004) 32(3) *Journal of Community Psychology* 229, 230-231.
[163] See e.g. *Statutory Declarations Regulations 2023* (Cth), Sch 1, r 1.
[164] Sung Hui Kim, 'Reimagining the Lawyer's Duty to Uphold the Rule of Law' (2023) 2023 *University of Illinois Law Review* 781, 810.

The most immediate implication of this emphasis in legal practice is again for a critical view of integrating AI into legal practice workflows. Yet it is also an opportunity for broader character formation of lawyers and law students towards practice cultures that place the good of others, and society as a whole, above career advancement or self-fulfilment. In practice, this may involve re-evaluating the role of pro bono work, to the extent that it has become 'an organizational imperative "institutionalized" within law firms'[166] or a reputation enhancer.[167]

Civic responsibility could be encouraged by emphasising legal clinic opportunities in practice and at law school,[168] integrating content on the roles of lawyers in broader society into compulsory papers like legal ethics, hosting regular seminars by lawyers serving marginalised communities (criminal defence, immigration, human rights, etc), or who are prominent in law reform campaigns, as part of professional development curricula, exploring 'shadowing' options for students with barristers or large law firms involved in pro bono work, involving lawyers and judges in legal education to a greater degree,[169] and adopting programs which involve law students educating the non-legally-qualified public about fundamental legal concepts.[170]

At first glance these options have nothing to do with AI. Perhaps that is the point. The verification-value paradox highlights the importance of truth because lawyers must be proven trustworthy. Trustworthiness can be built by inculcating in lawyers and law students a sense that as lawyers, they exist for others first, not themselves. This emphasis is consistent with the rules governing lawyer conduct – lawyers are charged with weighty duties to clients, the court, and others.

To that end, law is a high calling to *serve others first*. This attitude is certainly inconsistent with the cavalier adoption of AI-generated content in court submissions that has waylaid so many practitioners. The verification-value paradox suggests the solution is not in teaching lawyers, present and aspiring, how to wield AI effectively (even if that is possible). It is instead in cultivating the type of lawyers who will *not* make such mistakes precisely because they understand their role is to serve the administration of justice, the court and their clients, and that fidelity to the truth is vital to faithfully discharge that role.

---

[165] Ibid.
[166] Fiona Kay and Robert Granfield, 'When altruism is remunerated: Understanding the bases of voluntary public service among lawyers' (2022) 56 *Law & Society Review* 78, 94, citation omitted.
[167] Ching-fang Hsu, Ivan Kan-hsueh Chiang and Yun-chien Chang, 'Lawyers' legal aid participation: a qualitative and quantitative analysis' (2024) 21(2) *Journal of Empirical Legal Studies* 337, 364.
[168] Francina Cantatore and Nickolas J. James, 'Heroism Science Offers a New Framework for Cultivating Civic Virtue within Clinical Law Programs' (2017) 2 *Australian Journal of Clinical Education* [1], 6.
[169] Kari J. Kelso and J. Clark Kelso, 'Civic Education and Civil Discourse: A Role for Courts, Judges, and Lawyers' (2021) 21(2) *The Journal of Appellate Practice and Process* 473, 495-496.
[170] Anil Balan, 'Bridging the Gap: Law Students as Agents of Public Legal Education and Community Empowerment' (2025) 32(2) *International Journal of Clinical Legal Education* 53, 58-59.

# VI CONCLUSION

Discourse in favour of incorporating generative AI into legal practice workflows and legal education is based on a risk-opportunity paradigm. This paradigm suggests AI's risks can be tamed in favour of greater efficiency and efficacy in practice. However, it understates the considerable structural flaws of the particular types of AI the profession seeks to integrate: disconnection from facts and a lack of transparency.

This paper presented an alternative paradigm to evaluate the use of AI in legal practice: the verification-value paradox. This paradox suggests the gains AI is purported to bring to the profession are often overstated, because of the emphasis placed on lawyers to verify the accuracy of all content generated by AI. In a text-dominated profession, the greater the AI use, the greater the cost of manual verification. That imperative is clear from regulatory frameworks governing the conduct of lawyers, specific guidelines on the use of AI in court proceedings, and judicial criticism of lawyers who have presented unverified AI-generated material before the courts.

This paradox suggests lawyers current and future should treat AI with great scepticism to the extent it is purported to enhance legal practice, given manual verification costs. The paradox also encourages the cultivation of truth and civic responsibility in legal practice and education, developing lawyers future and current in alignment with their paramount ethical and professional obligations. Practical manifestations of these emphases could include secure assessment in law schools and encouraging community service and engagement by practitioners and law students, while ensuring lawyers and law students are adequately informed about the risks AI poses to the law and society more generally. Continuing professional development for lawyers will also take on increased importance as clients and other practitioners integrate AI into their workflows; a key skill will be for lawyers to distinguish between, and engage, AI-generated content in ethical and effective ways.

Of course, these recommendations are starting points. They require more thought for effective integration in legal practice and education, sensitive to context, jurisdictions, the makeup of the student body, and other factors.[171] Nevertheless, it is hoped that the analysis above refocuses lawyers and law schools on the high standards placed on them by lawmakers, regulators, and ultimately the public, allowing them to more critically appraise often illusory narratives about technology and progress while maintaining fidelity to the administration of justice, the court and the public.

---

[171] See e.g. Richard Wu and JaeWon Kim, 'An Empirical Study of Values of Law Students in South Korea: Does 'Americanized' Legal Education Impact Their Confucian Ethics?' (2022) 17 *University of Pennsylvania Asian Law Review* 209 on the integration of positive elements from US-style legal education with Confucian values influential in Korean law students.